\documentclass{article}
\usepackage{ijcai26}

\usepackage{times}
\usepackage{soul}
\usepackage{url}
\usepackage[hidelinks]{hyperref}
\usepackage[utf8]{inputenc}
\usepackage[small]{caption}
\usepackage{graphicx}
\usepackage{amsmath}
\usepackage{amsthm}
\usepackage{booktabs}
\usepackage{algorithm}
\usepackage{algorithmic}
\usepackage[switch]{lineno}
\usepackage[capitalise,noabbrev]{cleveref} 
\crefname{appsec}{Appendix}{Appendices} 
\usepackage{framed}
\usepackage{subcaption}
\usepackage{xcolor}
\usepackage{tikz}
\usetikzlibrary{arrows.meta}      %
\usetikzlibrary{shapes.geometric} %
\usetikzlibrary{positioning}      %
\usetikzlibrary{calc}             %
\usetikzlibrary{fit}              %
\usetikzlibrary{backgrounds}      %

\usepackage{enumitem}
\usepackage{comment}
\usepackage{tabularx}

\newtheorem{example}{Example}

\usepackage{nicefrac}

\newcommand{\blfootnote}[1]{%
  \begingroup
  \renewcommand\thefootnote{}\footnote{#1}
  \addtocounter{footnote}{-1}
  \endgroup
}

\title{CLPIPS: A Personalized Metric for AI-Generated Image Similarity}

\author{
Khoi Trinh,
Jay Rothenberger,
Scott Seidenberger,
Dimitrios Diochnos,
Anindya Maiti \\
\textnormal{University of Oklahoma} \\
\textnormal{\textit{khoitrinh@ou.edu, jay.c.rothenberger@ou.edu,\\seidenberger@ou.edu, diochnos@ou.edu, am@ou.edu}}
}

\definecolor{MyGreen}{cmyk}{0.95, 0.05, 0.95, 0.05}

\begin{document}

\maketitle

\begin{abstract}

Iterative prompt refinement is central to reproducing target images with text-to-image generative models. Previous studies have incorporated image similarity metrics (ISMs) as additional feedback to human users. These existing ISMs such as LPIPS and CLIP provide objective measures of image likeness but often fail to align with human judgments, particularly in context-specific or user-driven tasks. In this paper, we introduce Customized Learned Perceptual Image Patch Similarity (CLPIPS), a customized extension of LPIPS that adapts a metric's notion of similarity directly to human judgments. We aim to explore whether lightweight, human-augmented fine-tuning can meaningfully improve perceptual alignment, positioning similarity metrics as adaptive components for human-in-the-loop workflows with text-to-image tools. 
We evaluate CLPIPS on a human-subject dataset in which participants iteratively regenerate target images and rank generated outputs by perceived similarity. Using margin-ranking loss on human-ranked image pairs, we fine-tune only the LPIPS layer-combination weights and assess alignment via Spearman’s rank correlation %
and Intraclass Correlation Coefficient. %
Our results show that CLPIPS achieves stronger correlation and agreement with human judgments than baseline LPIPS. Rather than optimizing absolute metric performance, our work emphasizes improving alignment consistency between metric predictions and human ranks, demonstrating that even limited human-specific fine-tuning can meaningfully enhance perceptual alignment in human-in-the-loop text-to-image workflows.

\end{abstract}

\section{Introduction} \blfootnote{This study was reviewed and approved by the authors' Institutional Review Board (IRB).} 
\label{sec:intro}

Generative AI has made it possible to create images from text prompts, but reproducing a specific target image through prompt inference remains a challenge. In such tasks, users iteratively adjust prompts to better match a given reference image \cite{trinh2024promptly,trinh2025picture}. A key bottleneck is producing an image similarity metric that aligns with human perception. Existing image similarity metrics (ISMs) such as Learned Perceptual Image Patch Similarity (LPIPS) \cite{zhang2018perceptual} and CLIP-based scores \cite{clip} can provide feedback on image likeness. However, these ISMs' alignment with humans' subjective judgment remains unverified, especially in iterative workflows.

Iterative refinement has been a common strategy across many domains where humans progressively improve outputs using feedback, including writing, programming, and design. Flower~\shortcite{flower1981cognitive} provides a model on how writers plan, create, and revise their work iteratively. 
Madaan et al.~\shortcite{madaan2024self} show that iterative refinement is effective in significantly improving outputs in text and code generation tasks through self-feedback mechanisms. M{\o}ller and Aiello~\shortcite{moller2025prompt} show that stepwise prompt refinement can show improvement in text summarization tasks. Du et al.~\shortcite{du2022read} provides the R3 framework that has demonstrated the effectiveness of iterative revision in producing high-quality textual outputs by incorporating user feedback at each stage of the revision. 

Moreover, with the current rapid development of text-to-image models, the ability to iteratively reconstruct or approximate a target image using text prompts is relevant in several practical and economically meaningful contexts. These include but are not limited to: prompt recovery or prompt stealing, where users attempt to replicate proprietary or undisclosed prompts from observed outputs; bypassing prompt marketplaces or subscription gated prompt repositories by approximating high performing prompts through interaction alone; educational settings, where novice users learn how prompt structure maps to visual outcomes through guided feedback; creative restoration tasks, such as approximating lost, damaged, or incomplete visual artifacts when only partial references remain; and auditing and interpretability workflows, where understanding how prompts influence outputs helps diagnose model behavior.

Recent work by Trinh et al.~\shortcite{trinh2025picture} found that off-the-shelf ISMs only moderately reflect human judgment in similarity for prompt refinement tasks. When such metrics fail to reflect human perception, metric-guided refinement can become unreliable, as users may be encouraged to pursue prompt changes that improve numerical scores while degrading perceptual similarity, analogous to benchmark overfitting or metric gaming in broader machine learning contexts \cite{roth2024specification,hutchinson2022evaluation}. This gap motivates the need for more personalized or adapted metrics that better mirror human perception \cite{trinh2025picture}.

Rather than introducing yet another similarity metric optimized for raw accuracy or benchmark performance, in this paper, we focus on how well a metric's notion of similarity mirrors that of a human. To this end, we introduce \textbf{CLPIPS: Customized Learned Perceptual Image Patch Similarity}, a novel, fine-tuned ISM built upon LPIPS. CLPIPS is designed to better align with individual human judgments of image similarity by fine-tuning LPIPS using human generated ranking data. While LPIPS provides a strong baseline for perceptual distance, CLPIPS further adjusts this distance function using human similarity ranking. This calibration effectively guides the metric to what users consider important in the context of image comparison.

We evaluated CLPIPS on a dataset in which users iteratively generated prompts to match a target image and subsequently provided perceptual similarity rankings. Our evaluation utilizes two key metrics that quantify correlations between CLPIPS and human rankings: \textbf{Spearman's rank correlation coefficient ($\mathbf{\rho}$)} and the \textbf{Intraclass Correlation Coefficient }(ICC). We show that CLPIPS achieves a higher correlation with human similarity rankings than baseline LPIPS, with improvements that are statistically significant. Finally, we discuss how CLPIPS can lay a foundation for future on-the-fly personalization of similarity metrics during human-AI workflows.

Our work offers the following contributions.

\begin{enumerate}
    \item \textbf{Alignment-Oriented and Data-Efficient Similarity Metric:} We propose CLPIPS, a customized extension of LPIPS that is fine-tuned on human similarity judgments. CLPIPS employs a lightweight tuning strategy that calibrates LPIPS' deep features using human-ranked image data while updating only a small set of weights. To our knowledge, this is one of the first image similarity metrics explicitly adapted to individual/crowd preferences in an image regeneration context. For replicability purposes, the code and dataset will be made publicly available pending acceptance.
    \item \textbf{Evaluation of Human Judgments:} We evaluate CLPIPS against baseline LPIPS using a dataset of human similarity rankings derived from iterative image regeneration workflows. The dataset provides human similarity rankings over image sets, enabling robust evaluation of alignment via Spearman’s rank correlation and Intraclass Correlation Coefficient (ICC). We further assess the statistical significance and generalization of these improvements on held-out image sets.
    \item \textbf{Insights:} We demonstrate that even a modest amount of human-specific training data can noticeably improve metric alignment with subjective perception. We discuss the implications for interactive generative systems and outline how CLPIPS could enable on-the-fly metric adaptation to each user’s preferences in future applications. 
\end{enumerate}

\section{Background \& Related Work}
\label{sec:related}

\subsection{Learned Perceptual Image Patch Similarity and other ISMs}
\label{sec:ism}

Accurately quantifying visual similarity as humans perceive it is a long-standing challenge in computer vision. Traditional metrics like Euclidean Distance, structural similarity index measure
 (SSIM), and peak signal-to-noise ratio (PSNR) capture pixel-wise differences but often fall short in reflecting the perceptual nuances emphasized by human judgment. To address this, Zhang et al.~\shortcite{zhang2018perceptual} proposed using deep features from convolutional neural networks (CNNs). This approach resulted in Learned Perceptual Image Patch Similarity (LPIPS), which computes feature embeddings from multiple convolutional layers of pre-trained networks (e.g., AlexNet, VGG, or SqueezeNet), normalizes these activations channel-wise, and calculates the weighted L2 distance between feature maps. The final similarity score averages these distances across spatial dimensions and layers. LPIPS leverages the hierarchical nature of CNNs, allowing the metric to capture higher-order image structures and context-dependent visual patterns that impact human perceptual judgments. LPIPS has since been widely adopted as a perceptual similarity measure. Other modern approaches include multimodal models like CLIP, which compute similarity in a joint image-text embedding space \cite{clip}. The LPIPS and CLIP similarity metrics provide useful feedback in generative workflows, including iterative prompt refinement.

Despite their success, generic ISMs may not fully align with subjective human judgments in specific contexts. For example, Trinh et al.~\shortcite{trinh2025picture} found that in prompt-based image regeneration, existing metrics only moderately agree with human similarity rankings. This is unsurprising: out-of-the-box metrics are typically trained on general image datasets and approximate ``average" human perception rather than adapting to individual preferences or specific task domains. For example, LPIPS yields a single global similarity scale and does not account for which attributes (color, composition, semantic content, style) an individual user prioritizes when judging similarity. This misalignment can reduce trust in the metric if its notion of similarity diverges from the human’s.

\subsection{Quantifying Objective and Subjective Alignment}
\label{sec:alignment}

To quantitatively assess alignment between a metric and human judgments, we use reliability measures that quantify agreement and consistency between different evaluators. In particular, the Intraclass Correlation Coefficient (ICC) treats the metric as one ``rater'' and a human observer as another, measuring their consistency across many items \cite{koo2016guideline}. ICC accounts for both correlation and absolute agreement in ratings, providing a stringent test of alignment. We complement ICC in our analysis with Spearman’s rank correlation, which measures monotonic consistency in orderings.

Recent research has highlighted the limitations of existing similarity metrics and the need for stronger human alignment. Fu et al.\shortcite{fu2023dreamsim} introduced \textbf{DreamSim}, which learns new dimensions of visual similarity from synthetic data annotated with human judgments. DreamSim outperforms LPIPS in some semantic settings. Ghildyal et al.~\shortcite{ghildyal2025alignment} showed that \textbf{foundation model features} can boost low-level perceptual similarity metrics, yielding stronger agreement with humans.  Ghildyal’s dissertation provides a broader analysis of the robustness and human alignment of perceptual similarity metrics, including LPIPS \cite{ghildyal2025alignment}. Ahlert et al.\ conducted a systematic comparison of alignment metrics and found that correlations with human perception remain far from perfect, even across modern approaches \cite{ahlert2024aligned}. 

Overall, while perceptual similarity metrics have advanced, they still fall short of reproducing task-specific or user-specific human judgments. Our work addresses this gap by directly fine-tuning LPIPS on \emph{human-provided rankings from prompt-regeneration tasks}, focusing on whether the metric can reproduce human rank orderings rather than only correlating with raw values.

\subsection{Towards Fine-Tuned Perceptual Metrics}

Human judgment of similarity or aesthetics varies by individual. In personalized image similarity assessment, models are trained to predict user-specific preferences, using datasets with many participants and attributes. Pick-a-Pic introduces a public dataset of user comparisons and the PickScore model, which outperforms common baselines on predicting preferences and supports better ranking of generated images~\cite{kirstain2023pick}. Human Preference Score (HPS) fine-tuned CLIP on curated human preferences datasets and reports stronger agreement than baseline CLIP \cite{wu2023human}.

\subsection{Motivation}

A key departure from prior work on perceptual metrics is that our focus lies not on the magnitude of similarity scores but on their ordering consistency with human judgments. For perceptual comparison tasks such as prompt-based image regeneration, what matters is not the magnitude of the distance or metric itself but whether the distance or metric ranks images in the same order that humans do. We therefore treat the metric as a ranking function and evaluate it using ordinal reliability measures (Spearman’s $\rho$ and ICC) that capture how faithfully the model reproduces human rank orderings. This emphasis on rank-level alignment shifts the goal of similarity modeling from value prediction to preference reproduction, aligning more closely with how humans naturally judge ``which of two images looks closer" rather than ``how similar is this on a 0–1 scale."

\section{Research Questions \& Hypotheses}
\label{sec:goals}

Building on previous work by Trinh et al.~\shortcite{trinh2025picture}, highlighting the gap between objective image similarity metrics and subjective human perception, this study investigates whether fine-tuning perceptual metrics (specifically LPIPS) using human similarity ranking data can yield better-aligned outcomes. We formalize our research through the following research questions and hypotheses.

\paragraph{RQ1: Does fine-tuning LPIPS with rank-based data improve the alignment consistency between metric predictions and human judgments?}

First, CLPIPS aims to mimic human judgment by introducing fine-tuning with user ranking data in order to improve on existing ISMs such as LPIPS \cite{zhang2018perceptual} or CLIP \cite{clip} which only exhibit moderate alignment with human ranking \cite{trinh2025picture}. Prior studies have also brought forth the need for metrics that adapt to user-specific or task-specific nuances \cite{trinh2025picture,trinh2024promptly,ahlert2024aligned,ghildyal2025alignment}. Our hypotheses for RQ1 are as follows:

\begin{itemize}[noitemsep, topsep=0pt, left=0pt]
    \item \textbf{Hypothesis 1.1:} CLPIPS will achieve higher Spearman's rank correlation coefficient $\rho$ compared to baseline LPIPS because it is aligned more closely with user preferences.
    \item \textbf{Hypothesis 1.2:} CLPIPS will achieve higher Intraclass Correlation Coefficient (ICC) than baseline LPIPS, indicating stronger consistency with human orderings.
\end{itemize}

\paragraph{RQ2: Is this improvement statistically and practically meaningful?}

While personalization may increase alignment numerically, it is essential to assess whether these gains are statistically reliable. Statistical tests of correlation and significance are therefore used to quantify significance \cite{koo2016guideline,cicchetti1994guidelines}. Our hypotheses for RQ2 are as follows:

\begin{itemize}[noitemsep, topsep=0pt, left=0pt]
    \item \textbf{Hypothesis 2.1:} The improvement of CLPIPS over LPIPS for both Spearman's rank correlation $\rho$ and Intraclass Correlation Coefficient (ICC) will be statistically significant ($p<0.05$).
    \item \textbf{Hypothesis 2.2:} The observed ICC improvement will shift the alignment toward a higher reliability category, following established guidelines.
    \item \textbf{Hypothesis 2.3:} The observed ICC improvement will not be driven by a small number of target images.
\end{itemize}

These tests and metrics are further discussed in \cref{sec:eval}.

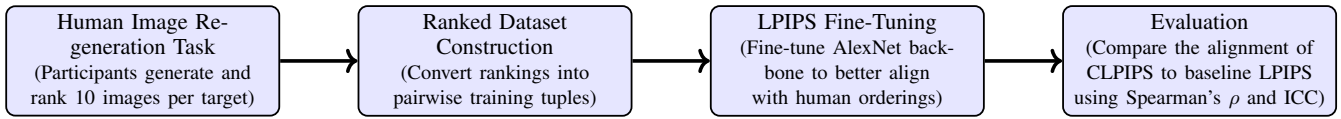
\begin{figure*}[t]
  \centering
  \resizebox{\textwidth}{!}{%
  \begin{tikzpicture}[
      node distance=1.6cm and 1.2cm,
      process/.style={rectangle, rounded corners, draw=black, thick, fill=blue!10, text width=4cm, align=center, minimum height=1.2cm},
      arrow/.style={->, ultra thick}
    ]

    \node[process] (human) {Human Image Regeneration Task\\
    \footnotesize (Participants generate and rank 10 images per target)};
    
    \node[process, right=of human] (dataset) {Ranked Dataset Construction\\
    \footnotesize (Convert rankings into pairwise training tuples)};
    
    \node[process, right=of dataset] (training) {LPIPS Fine-Tuning\\
    \footnotesize (Fine-tune AlexNet backbone to better align with human orderings)};
    
    \node[process, right=of training] (eval) {Evaluation\\
    \footnotesize (Compare the alignment of CLPIPS to baseline LPIPS using Spearman’s $\rho$ and ICC)};

    \draw[arrow] (human) -- (dataset);
    \draw[arrow] (dataset) -- (training);
    \draw[arrow] (training) -- (eval);
  \end{tikzpicture}
  }%
  \caption{Workflow of CLPIPS fine-tuning. Human-generated and ranked images are converted into pairwise tuples to fine-tune LPIPS using a margin ranking loss, producing CLPIPS, which is then evaluated against human similarity rankings.}
  \label{fig:clpips_workflow}
\end{figure*}

\begin{figure*}[htbp]
\centering
\begin{subfigure}[b]{0.4\linewidth}
    \centering
    \includegraphics[width=\textwidth]{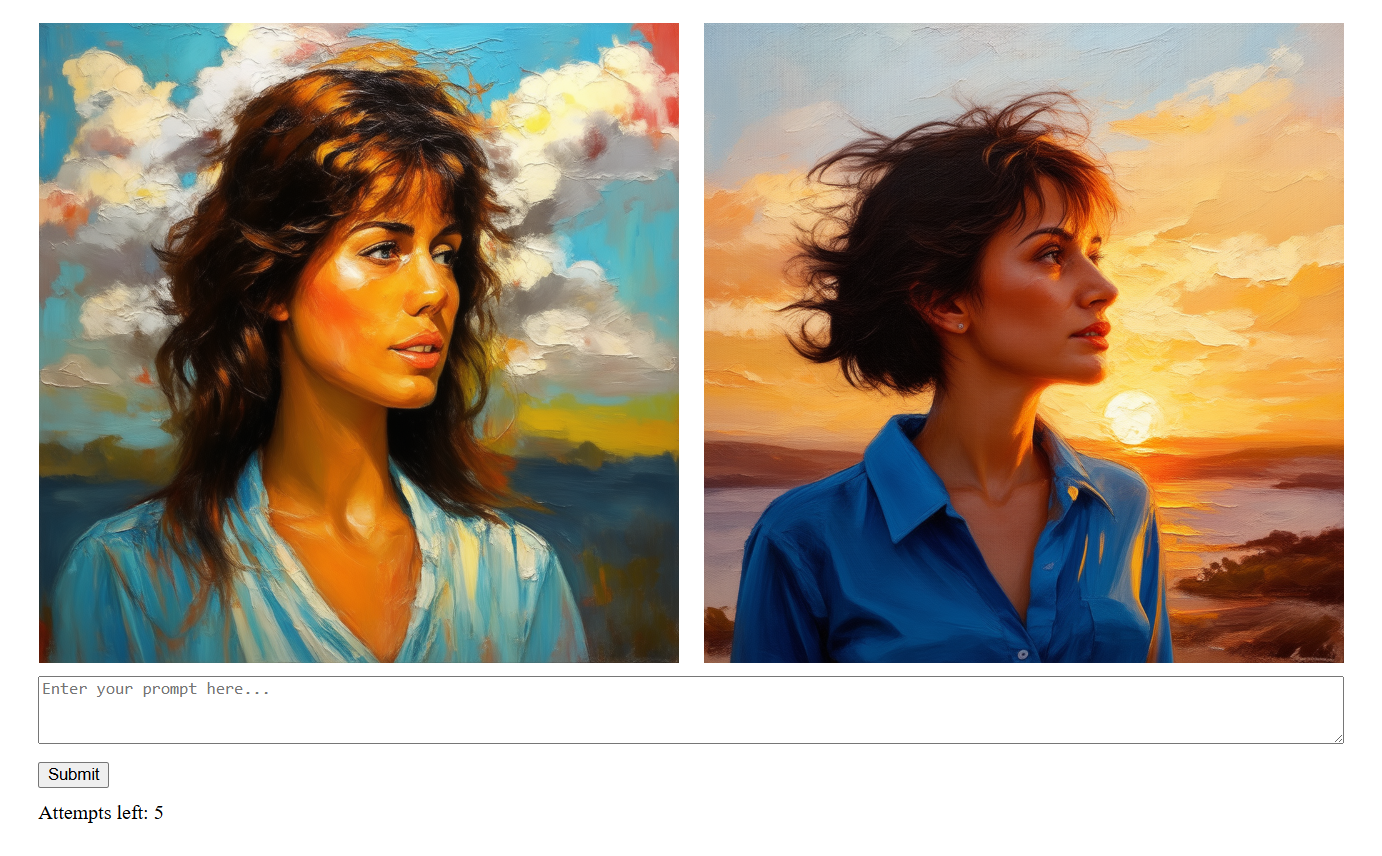}
    \caption{Task: Iterative Prompt Refinement}
    \label{fig:iter-task}
\end{subfigure}
\quad
\begin{subfigure}[b]{0.5\linewidth}
    \centering
    \includegraphics[width=\textwidth]{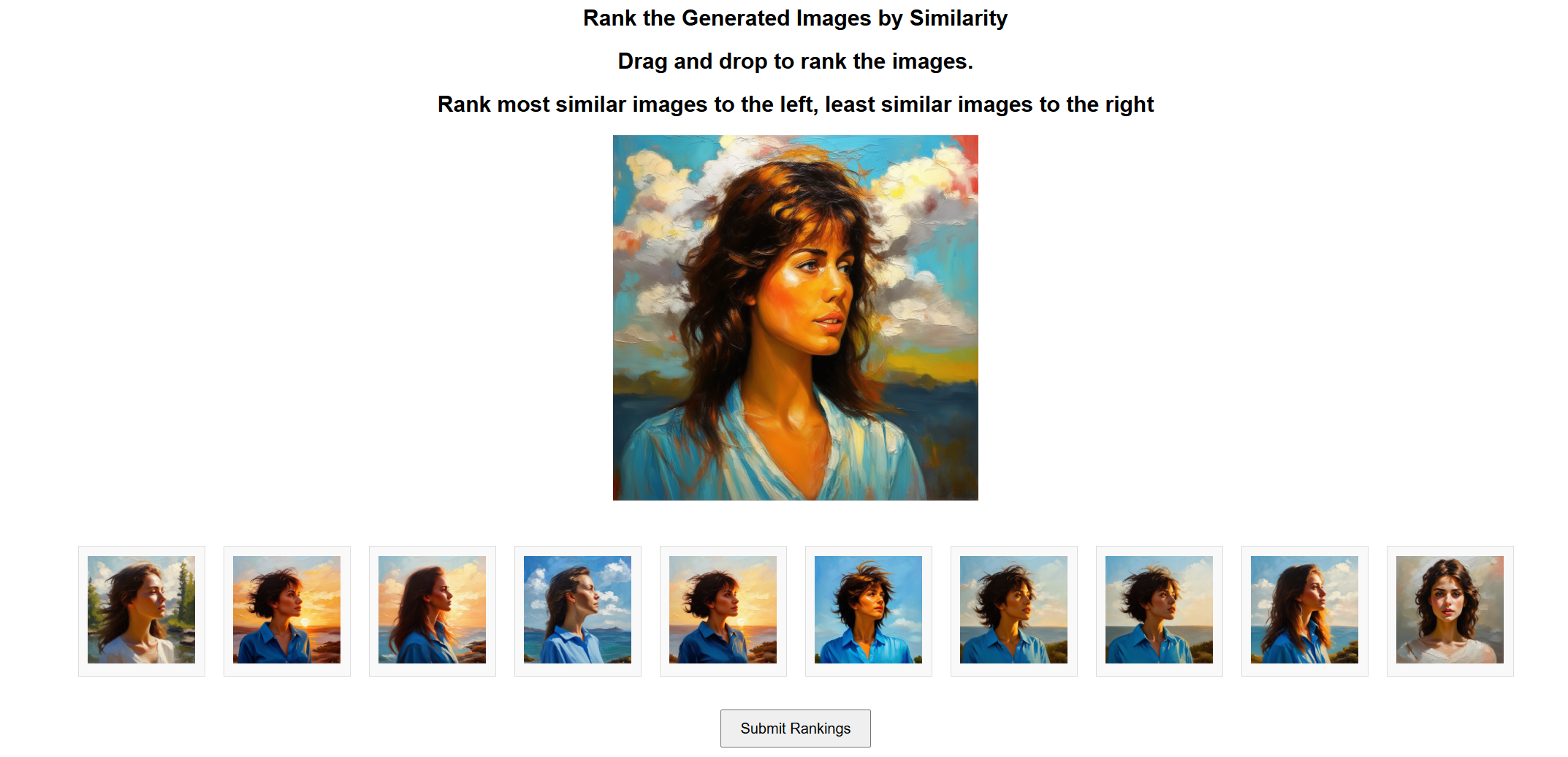}
    \caption{Task: Subjective Similarity Ranking}
    \label{fig:ranking-task}
\end{subfigure}
\caption{An example of the two main survey tasks.}
\label{fig:demographics}    
\end{figure*}

\section{Methodology and Experimental Setup}
\label{sec:exp-setup}

\subsection{Overview}

The goal of CLPIPS is to personalize the LPIPS metric using human similarity judgments. We frame this as a supervised metric tuning problem: given a target image and a set of 10 generated images, we want the learned metric to predict which generated image is more similar to the target according to the human. We accomplish this by fine-tuning the weights of LPIPS on a set of labeled image pairs derived from human rankings. \cref{fig:clpips_workflow} provides a high-level outline: we take the pre-trained weights from the AlexNet backbone of LPIPS and adjust its linear combination layers using a custom training dataset of ``more similar to least similar'' image comparison.

\subsection{Survey Design and the Dataset}

\subsubsection{Task: Image Regeneration through Iterative Prompt Refinement}
\label{sec:survey-main-task-1}

To create our human-ranking dataset, we conducted a survey with 20 participants. Each participant was assigned a set of 10 target images, and tasked with refining prompts over 10 iterations per image to reproduce the target image. The participant pool of 20 was divided into four groups of five, each working with a distinct subset of target images to ensure balanced coverage across the dataset. \cref{fig:iter-task} shows an example of this task.
 
\subsubsection{Task: Subjective Similarity Ranking}
\label{sec:survey-main-task-2}

After completing the iterative refinement task, participants ranked the 10 images generated across iterations for each target image. Rankings were performed using a drag-and-drop interface, placing the most visually similar images on the left and the least similar on the right. \cref{fig:ranking-task} shows an example of this task.

\subsection{Model Architecture}

The primary baseline in our evaluation is LPIPS with the AlexNet backbone and its default learned weights from Zhang et al. \cite{zhang2018perceptual}, which represents a non-personalized perceptual similarity metric. We computed LPIPS distances for all target–image pairs in the test set and evaluated alignment with human rankings using the same Spearman and ICC measures applied to CLPIPS. This enables a direct comparison and quantification of the gains introduced by personalization. While we considered other metrics (e.g., CLIP image embedding cosine similarity) for contextual reference, our focus is on demonstrating improvement over the baseline upon which CLPIPS is built; accordingly, all reported correlations and statistical tests are limited to LPIPS and CLPIPS.

CLPIPS is constructed by fine-tuning the LPIPS framework while retaining its AlexNet backbone. LPIPS computes a distance $d_{\text{LPIPS}}(I_{\text{ref}}, I)$ by comparing deep feature representations across multiple layers and combining these layer-wise distances using a small set of learned linear weights. In CLPIPS, we freeze all convolutional layers to preserve general-purpose visual features and update only the layer-combination weights. This design drastically reduces the number of trainable parameters and mitigates overfitting, which is especially important given the modest size of our dataset. Conceptually, this approach assumes that the pre-trained deep features are sufficiently expressive, while allowing the relative importance of different visual attributes (e.g., color, texture, or structural details) to be adjusted to better reflect human similarity judgments.

\subsubsection{Loss Function:}

We train CLPIPS using a margin ranking loss (hinge loss) to encourage the metric to assign a smaller distance to the more-similar image and a larger distance to the less-similar image (by a margin $m$). For a triplet $(I_{tgt}, I_{pos}, I_{neg})$ where $I_{pos}$ is judged more similar to the target than $I_{neg}$, the loss is:

\begin{equation}
\mathcal{L} = \max\left\{0,\ d(I_{\text{tgt}}, I_{\text{pos}}) - d(I_{\text{tgt}}, I_{\text{neg}}) + m\right\}
\end{equation}
where $d(\cdot,\cdot)$ is the metric's predicted distance, computed as the weighted L2 distances between deep feature representations extracted from multiple convolutional layers \cite{zhang2018perceptual}.
We set a small margin (e.g., $m=0.03$) to only penalize violations where the ``better” image isn’t sufficiently closer than the ``worse” image. Minimizing this loss pushes CLPIPS to order pairs correctly. Fine-tuning was performed using a 70/30 split on the dataset, using 70\% of the target-image sets for training and holding out the remaining 30\% for validation to monitor covergence and prevent overfitting. The splits were defined at the level of entire target-image sets to ensure no overlap. A modest batch size (256 pairs) and Adam optimizer (learning rate $4\times10^{-4}$) were used.

\section{Evaluation Metrics}
\label{sec:eval}

We employed two complementary evaluation metrics: Spearman's rank correlation coefficients ($\rho$) and Intraclass Correlation Coefficients (ICC).

\subsection{Spearman's Rank Correlation Coefficient}
This correlation measures the monotonic agreement between two ranked variables. With an absence of ties in ranking, Spearman~\shortcite{spearman1904} defines the formula for this correlation as:
\begin{equation}\label{eq:rho}
\rho = 1 - \frac{6 \sum_{i=1}^{n} d_i^2}{n(n^2 - 1)}\,,
\end{equation}
where $n$ is the number of target images, and $d_i$ denotes the difference between the human-assigned rank and the ISM-assigned rank for image $i$. 

For each target image set, we compute the CLPIPS distance for each of the 10 generated images relative to the target. We then assess how well these distances correlate (inversely) with the human’s rank ordering. A high Spearman $\rho$ (close to 1.0 in absolute value) means that if the human ranks image A higher (more similar) than image B, CLPIPS also assigns a lower distance (higher similarity) to A than B, for all pairs. The results are aggregated over all image sets in the training data. In practice, we merged all (human rank, metric score) pairs across the test sets and computed Spearman’s $\rho$ on this aggregate (with appropriate tie-breaking for identical ranks). We also report the p-value for the null hypothesis of $\rho = 0$ to test for statistical significance.

\subsection{Intraclass Correlation Coefficient}

This correlation provides a more stringent test by measuring absolute agreement in rankings between CLPIPS and the human. We use a two-way random effects ICC, treating each set of images as a target ``item” and the human and metric as two ``raters” who each assign a rank (1–10) to each image in that set. The ICC reflects both consistency and agreement: it will be high only if, for each image set, the CLPIPS assigned ranks of the images closely match the human’s ranks.

Shrout and Fleiss~\shortcite{shrout1979intraclass} define the formula for a two-way, random effects ICC model measuring absolute agreement among multiple raters as:
\begin{equation}\label{eq:ICC}
\mathrm{ICC}(2,k) = 
\frac{MS_R - MS_E}
{MS_R + (MS_C - MS_E)/n}\,,
\end{equation}
where $n$ the number of target images, $k$ is the number of raters (here, $k = 2$, one for the human baseline, and one for the metric), 
$MS_R$ denotes the between-image ranking variance, $MS_C$ denotes the between-rater variability, and $MS_E$ is the error variability.

Koo and Li~\shortcite{koo2016guideline} define interpretive guidelines for ICC values, summarized in \Cref{tab:icc-baselines-values}. Cicchetti~\shortcite{cicchetti1994guidelines} provides an alternative set of thresholds, shown in \Cref{tab:icc-baselines-values-alt}.

\begin{table}[htbp]
\centering
\begin{tabular}{ll}
\toprule
\textbf{ICC Range} & \textbf{Interpretation} \\
\midrule
$< 0.50$ & Poor \\
$0.50$ -- $0.75$ & Moderate \\
$0.75$ -- $0.90$ & Good \\
$> 0.90$ & Excellent \\
\bottomrule
\end{tabular}
\caption{Interpretation of Intraclass Correlation Coefficient (ICC) values by Koo and Li \protect\shortcite{koo2016guideline}}
\label{tab:icc-baselines-values}
\end{table}

\begin{table}[htbp]
\centering
\begin{tabular}{ll}
\toprule
\textbf{ICC Range} & \textbf{Interpretation} \\
\midrule
$< 0.40$ & Poor \\
$0.4$ -- $0.6$ & Fair \\
$0.6$ -- $0.75$ & Good \\
$> 0.75$ & Excellent \\
\bottomrule
\end{tabular}
\caption{Interpretation of Intraclass Correlation Coefficient (ICC) values by Cicchetti \protect\shortcite{cicchetti1994guidelines}}
\label{tab:icc-baselines-values-alt}
\end{table}

\begin{example}\label{example:metrics}
Let the relation $I_1 \succ I_2$ denote that image $I_1$ is judged more similar to the target image than image $I_2$. A ranking over $n$ images is represented as a tuple of size $n$, which is an integral rank of $1$ to $n$, where rank $1$ corresponds to the most similar image (leftmost in the tuple) and rank $n$ corresponds to the least similar image (rightmost in the tuple).

Consider four images $A, B, C, D$. Let a baseline ranking $BR$ indicate the ideal ranking, and two other rankings $R1$ and $R2$, among these four image:
\begin{displaymath}
\left\{
\begin{array}{rcl}
    BR  &=& (A, B, C, D) \\
    R_1 &=& (A, D, B, C) \\
    R_2 &=& (D, B, A, C)
\end{array}
\right.
\end{displaymath} 
We see that $R_1$ suggests $A \succ D \succ B \succ C$;
with $\mathrm{rank}_{R_1}(A) = 1$, $\mathrm{rank}_{R_1}(B) = 3$, $\mathrm{rank}_{R_1}(C) = 4$, and $\mathrm{rank}_{R_1}(D) = 2$. Intuitively, $R1$ should be better aligned with $BR$ than $R2$.
Then, both Spearman's $\rho$ and ICC suggest that ranking $R_1$ is better aligned with $BR$ compared to $R_2$.
\end{example}

\begin{proof}[Justification Sketch for Example~\ref{example:metrics}]
We have the following.

\noindent\underline{Spearman's Rank Correlation Coefficient:}
Between $BR$ and $R_1$, the rank differences yield $\sum_{i=1}^4 d_i^2 = 0^2 + (-2)^2 + 1^2 + 1^2 = 6$. 
By (\ref{eq:rho}), 
$\rho_{R_1} = 1 - \frac{6\cdot 6}{4\cdot 15} = 1 - \nicefrac{36}{60} = \nicefrac{24}{60} = 0.4$

Similarly, between $BR$ and $R_2$, we have $\sum_i d_i^2 = %
14$. 
Hence, by (\ref{eq:rho}) now we have 
$\rho_{R_2} = 1 - \frac{6\cdot 14}{4\cdot 15} 
= \nicefrac{-24}{60} = -0.4$.

\smallskip

\noindent\underline{Intraclass Correlation Coefficient:}
We use the fact that $k=2$ raters are being used for every alignment comparison.

Regarding $BR$ and $R_1$ we have the following.
$MS_R = \frac{\sum_{i=1}^n 2(\overline{\mathrm{rank}(I_i)} - \overline{\mathrm{rank}})^2}{n-1} 
= \frac{7}{3}$.
Moreover, $MS_C = 0$ when we treat rankings as scores (as we do) since this involves summation of difference between the average rank of a rater and the average rank among all raters, which is 2.5 in every case.
Similar calculations show that $MS_E = 1$ by using the variances of the rankings of the individual images.  
Plugging these values into~(\ref{eq:ICC}) we obtain:
$\mathrm{ICC}_{R_1}(2, 2) 
= 0.64$,
which indicates ``Moderate" agreement between $R_1$ and $BR$, per~\cref{tab:icc-baselines-values}.

For $BR$ and $R_2$, we have $MS_R = 1$, $MS_C = 0$, and $MS_E = 2.33$, resulting in
$\mathrm{ICC}_{R_2}(2, 2) \approx -3.2$,
which is ``Poor" agreement between $R_2$ and $BR$, per \cref{tab:icc-baselines-values}.

For further details on the omitted ICC calculations, we refer interested readers to~\cite{ICC:Tutorial}.
\end{proof}

\section{Experimental Results}
\label{sec:results}
\subsection{RQ1: Correlation with Human Judgments}

CLPIPS showed a clear improvement over LPIPS in alignment with human image similarity judgments. When calculating Spearman’s rank correlation, we found that CLPIPS achieved a $\rho = 0.524$ with human rankings, compared to $\rho = 0.432$ for baseline LPIPS. This means that CLPIPS’s similarity scores had a stronger monotonic relationship with how humans ordered images by similarity.

\begin{figure}[htbp]
\centering
\includegraphics[width=\linewidth]{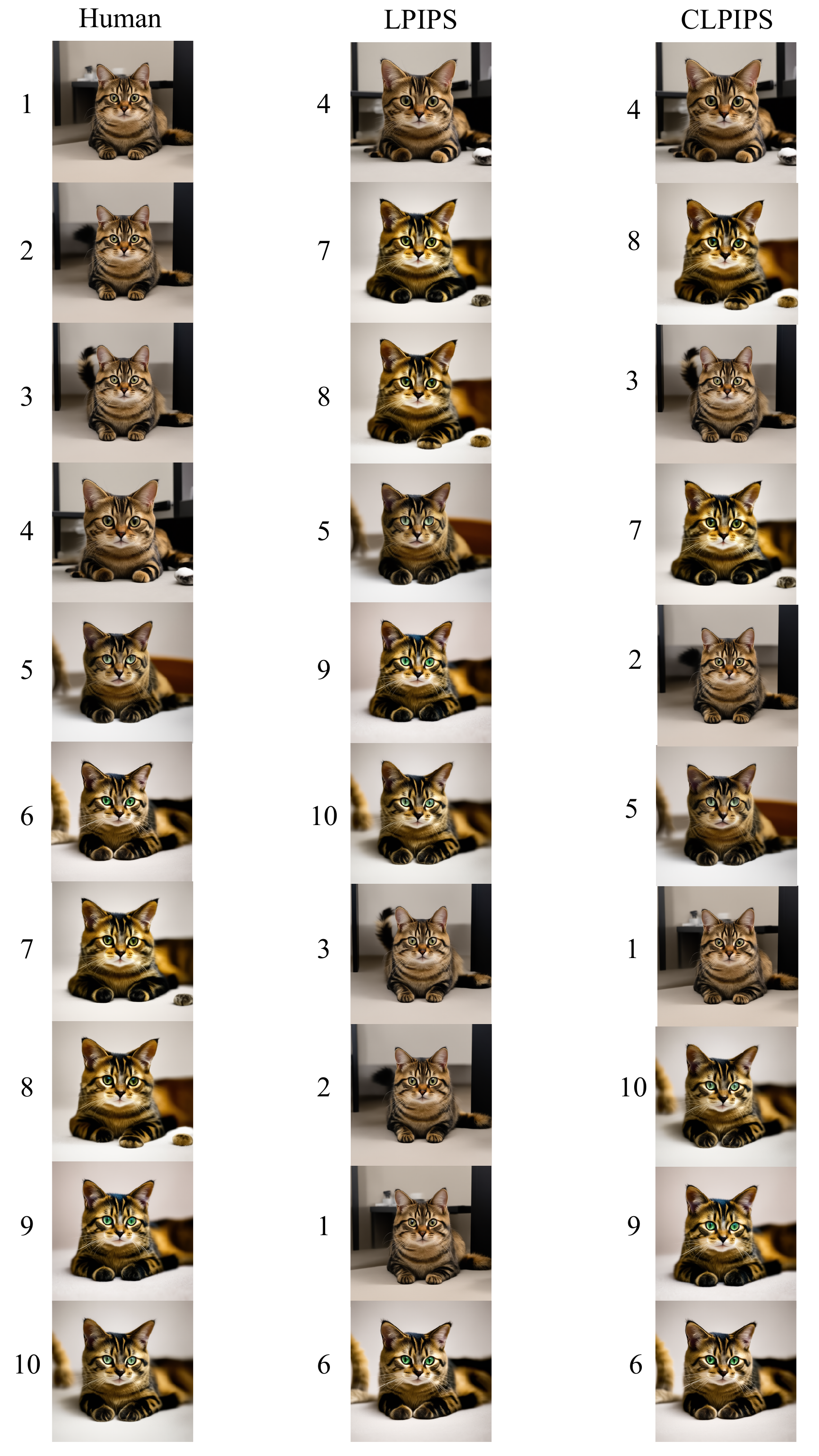} %
\caption{Comparison of human similarity rankings with LPIPS and CLPIPS rankings for the same set of generated images relative to a target image. Images are indexed according to their human-assigned rank (shown in the left column). The same index numbers are shown next to the corresponding images in the LPIPS and CLPIPS columns, indicating where each human-ranked image appears in the metric-based orderings. CLPIPS places images closer to their human-ranked positions than LPIPS, exhibiting fewer rank inversions. In this example, CLPIPS achieved an ICC of $0.45$ while LPIPS achieved an ICC of $-1.41$.
} 
\label{fig:ranking-sum}
\end{figure}

Next, when looking at ICC values, baseline LPIPS yielded an $ICC(2,k) = 0.60$, indicating ``moderate'' agreement. In contrast, CLPIPS achieved an $ICC(2,k) = 0.68$, which still remains the ``moderate" range, according to Koo and Li~\shortcite{koo2016guideline}. Using a different set of guidelines from Cicchetti~\shortcite{cicchetti1994guidelines}, LPIPS is in the ``fair" range while CLPIS achieved ``good" alignment. Regardless of the guidelines, this increase from $0.60$ to $0.68$ represented a clear improvement in the consistency with which the metric reproduced the order given by the human rater. \Cref{tab:alignment-results} shows these values, along with their associated $p$ values. Moreover, \Cref{fig:ranking-sum} provides a comparison of human rankings, LPIPS rankings, and CLPIPS rankings for a generated image set, illustrating how the improved metric–human alignment manifests itself at the level of individual image orderings.

\begin{table}[htbp]
\centering
\begin{tabular}{lcccccc}
\toprule
\textbf{Model} & \textbf{ICC$_{2k}$} & \textbf{95\% CI} & \textbf{$\mathbf{p}$-value} \\
\midrule
CLPIPS & 0.68 & [0.66, 0.71] & $\ll 0.001$ \\
LPIPS & 0.60 & [0.57, 0.64] & $\ll 0.001$ \\
\midrule
\multicolumn{7}{l}{\textit{Spearman rank correlation ($\rho$) with human ranks}} \\
\midrule
CLPIPS & \multicolumn{2}{c}{$\rho = 0.524$} & \multicolumn{3}{c}{} & $p \ll 0.001$ \\
LPIPS & \multicolumn{2}{c}{$\rho = 0.432$} & \multicolumn{3}{c}{} & $p \ll 0.001$ \\
\bottomrule
\end{tabular}
\caption{Alignment of metric rankings with human rankings.}
\label{tab:alignment-results}
\end{table}

\subsection{RQ2: Statistical Significance And Dataset Robustness}

Both the ICC and Spearman's $\rho$ analyses indicate that the model's improvement is statistically robust. We obtained $p$ values below $0.001$ for both ICC and Spearman's $\rho$; allowing us to reject the null hypothesis of no improvement. 

The corresponding 95\% confidence intervals (CIs) are relatively narrow, indicating low variance across target image sets and suggesting that CLPIPS’s gains are consistent across resampled target image sets. Moreover, to directly assess the robustness of the improvement in alignment, we conducted a paired bootstrap analysis over target image sets (\Cref{tab:bootstrap-icc}). The results shows that CLPIPS achieves a statistically significant improvement in agreement over LPIPS. Overall, these findings demonstrate that the improvement was statistically significant and reasonably generalizable, showing that CLPIPS captures human-perceptual preference rather than overfitting to the training images.

\begin{table}[htbp]
\centering
\begin{tabular}{lccc}
\toprule
\textbf{Comparison} & \textbf{$\Delta$ICC} & \textbf{95\% CI} & \textbf{$\mathbf{p}$-value} \\
\midrule
CLPIPS -- LPIPS & 0.084 & [0.071, 0.1] & $<< 0.001$ \\
\bottomrule
\end{tabular}
\caption{Paired bootstrap analysis comparing ICC agreement between CLPIPS and LPIPS over target image sets. $\Delta$ICC shows the average increase in agreement in CLPIPS relative to LPIPS, while the bootstrap confidence interval indicates the stability of this improvement across resampled image sets.}
\label{tab:bootstrap-icc}
\end{table}

\section{Discussion}
\label{sec:discussion}

The above results have important implications for both the design of human-AI systems and future research on personalized metrics. We discuss key observations and potential extensions.

\subsection{Hypothesis Evaluation}

The results shown in \Cref{tab:alignment-results} allow us to evaluate each hypothesis introduced in \Cref{sec:goals}. The outcomes are summarized as follows:

\begin{itemize}
    \item \textbf{H1.1 (Spearman's rank correlation):} \textit{Supported.} CLPIPS achieved a higher $\rho$ (0.524 vs 0.432), indicating stronger monotonic agreement with human similarity rankings.
    \item \textbf{H1.2 (ICC improvement):} \textit{Supported.} CLPIPS achieved an ICC$_{2k}$ of 0.68, exceeding LPIPS's 0.60, indicating improved consistency in reproducing human rank orderings. 
    \item \textbf{H2.1 (Statistical significance):} \textit{Supported.} All tests yield extremely small $p$-values ($p \ll 0.001$), rejecting the null hypothesis of no alignment improvement and demonstrating that CLPIPS’s improvements are statistically robust.
    \item \textbf{H2.2 (Alignment category shift):} \textit{Partially supported.} Although CLPIPS demonstrated improved ICC ($0.68$ compared to LPIPS' ICC of $0.60$), the shift in alignment category depends on which guideline is used. Per Koo and Li in \Cref{tab:icc-baselines-values}, both LPIPS and CLPIPS achieved ``good" alignment agreement. Howwever, using Cicchetti's guidelines in \Cref{tab:icc-baselines-values-alt}, CLPIPS was able to move into the ``good" category, from ``fair" for LPIPS.
    \item \textbf{H2.3 (Robustness across dataset):} \textit{Supported.} A paired bootstrap analysis over target image sets shows that CLPIPS consistently achieves better ranking alignment than LPIPS, with a positive $\Delta ICC$ observed across all bootstrap resamples. This indicates that our observed improvement from fine-tuning is stable across the dataset, and not driven by only a small number of target images.
\end{itemize}

These evaluations collectively confirm that CLPIPS achieves statistically significant improvements in alignment with human similarity rankings while maintaining moderate-to-good reliability under established guidelines \cite{koo2016guideline}.

\subsection{Implications from Results}
\label{sec:implications}

Our findings confirm that the default LPIPS, despite being ``perceptual”, was not fully aligned with the specific notion of similarity in our task, and had room for improvement \cite{trinh2025picture}. Moreover, while only fine-tuned on a relatively small dataset of 2000 image sets, CLPIPS was able to achieve a notable shift in mimicking human ranking preference, effectively learning the nuances that humans cared about in the context of iterative prompt refinement. This suggests that certain visual features or differences were consistently either under-weighted or over-weighted by LPIPS, relative to human judgment, and the training process adjusted for that. For example, it is possible that LPIPS, which was trained using generic image distortions and patches \cite{zhang2018perceptual}, might emphasize low-level texture differences that humans ignored in our task, while humans could place more weight on semantic content alignment or style coherence between the target and generated image. CLPIPS likely down-weighted some of those texture-sensitive features and up-weighted higher-level similarities.

\subsection{Limitations and Future Work}
\label{sec:limitations-futurework}

While CLPIPS improves alignment, it is not perfect. An ICC of 0.55 still leaves room for improvement in order to approach near-human agreement. In part, this may be due to noise and variability in human judgment. There may be an upper bound on how high a single metric can align with one human or a group of humans. Additionally, our approach currently creates a single personalized model from the entire training set. We did not explore dynamic per-user adaptation within the study. It’s possible that certain users had systematic preferences that differ from others (e.g., some might focus more on color matching, others on object shape). A single model can only capture the average of these preferences.

Moreover, this study relies on a controlled human-subject dataset collected for a specific prompt refinement task. Although the dataset is limited in scale, the stability of the bootstrap analysis (refer to the 95\% CI in \Cref{tab:bootstrap-icc}) suggests that it is sufficient enough to support a meaningful and consistent comparison. While this dataset provides structured, high-quality ranking supervision, larger and more diverse datasets spanning additional visual domains, user populations, and task contexts may enable stronger generalization and further improvements in alignment. %
Future work could explore scaling this approach to broader datasets as well as integrating CLPIPS into real-time or on-the-fly fine-tuning during live user interaction. This approach of online personalization could reduce reliance on offline, static dataset and allow CLPIPS to adapt to user preference within their image regeneration workflows.

Lastly, while CLPIPS demonstrated improved alignment between metric and human ranking, this setting has not directly tested whether the new feature weights transfer to new, unseen images. Assessing generalization is important when deploying in open-ended text-to-image workflows (e.g., on-the-fly or real-time fine-tuning as previously mentioned). Future work could address this by introducing evaluation of CLPIPS on unseen, held-out images, such as those from publicly available dataset of images.

Such investigations would help determine the broader applicability of iterative prompt refinement across generative AI modalities.

\section*{Conclusion}
\label{sec:conclusion}

We presented CLPIPS, a personalized image similarity metric that fine-tunes the popular LPIPS metric using human similarity judgments. Our comprehensive evaluation shows that CLPIPS outperforms the baseline LPIPS in aligning with human perceptions of image similarity, as evidenced by higher Spearman correlations and ICC values with respect to human rankings of generated images. CLPIPS serves as a proof-of-concept that even lightweight personalization, adjusting a pre-trained metric with a relatively small amount of user data, can yield meaningful improvements in how well an automated metric reflects subjective human criteria.

\bibliographystyle{named}
\bibliography{references}

\appendix

\end{document}